\documentclass[11pt]{article}
\usepackage{coling2018}
\usepackage{times}
\usepackage{latexsym}

\usepackage{url}

\usepackage{extarrows}
\usepackage{amssymb,amsfonts}
\usepackage{latexsym}
\usepackage{multirow}
\usepackage{arydshln}
\usepackage{amsmath}
\usepackage{rotating}
\usepackage{color}
\usepackage{subfigure}

\def\figref#1{Figure~\ref{fig:#1}}
\def\figlabel#1{\label{fig:#1}\label{p:#1}}

\def\secref#1{\S\ref{sec:#1}}
\def\seclabel#1{\label{sec:#1}\label{p:#1}}
\def\eqref#1{Eq \ref{eqn:#1}}
\def\eqsref#1#2{Eqs.~\ref{eqn:#1}--\ref{eqn:#2}}
\def\eqlabel#1{\label{eqn:#1}}

\def\dnrm#1{\mbox{$_{\hbox{\scriptsize #1}}$}}

\long\def\eat#1{\ignorespaces}

\title{Recurrent One-Hop Predictions  \\ for Reasoning over Knowledge Graphs}

\author{Wenpeng Yin$^1$, Yadollah Yaghoobzadeh$^2$, Hinrich Sch\"{u}tze$^3$\\
  $^1$Dept.\ of Computer Science, University of
  Pennsylvania, USA\\
  $^2$ Microsoft Research, Montreal, Canada \\
  $^3$CIS, LMU Munich, Germany\\
  {\tt
    wenpeng@seas.upenn.edu}
   }

\date{}

\newcounter{notecounter}
\newcommand{\enotesoff}{\long\gdef\enote##1##2{}}
\newcommand{\enoteson}{\long\gdef\enote##1##2{{
\stepcounter{notecounter}
\large\bf
\hspace{1cm}\arabic{notecounter} $<<<$ ##1: ##2
$>>>$\hspace{1cm}}}}
\enoteson
\enotesoff

\newcommand{\modelname}{ROP}

\begin{document}
\maketitle
\begin{abstract}
Large scale knowledge graphs (KGs) such as Freebase are
generally incomplete.  Reasoning over
multi-hop (mh) KG paths is thus an important capability that is needed for
question answering or other NLP tasks that require knowledge
about the world.  mh-KG reasoning includes diverse scenarios,
e.g., given a head entity and a relation path, predict the
tail entity; or given two entities connected by some
relation paths, predict the unknown relation between them.
We present \emph{\modelname{}s}, recurrent one-hop predictors, that
predict entities at each step of mh-KB paths by using recurrent neural networks and  vector representations of entities and relations, with two
benefits: (i) modeling mh-paths of arbitrary lengths while
updating the entity and relation representations by the
training signal at each step; (ii) handling different types
of mh-KG reasoning in a unified framework.  Our models show state-of-the-art for two important multi-hop KG reasoning tasks:
Knowledge Base Completion and Path Query Answering.\footnote{\url{https://github.com/yinwenpeng/KBPath}}
\end{abstract}


\section{Introduction}\label{sec:intro}

\blfootnote{
    %
    %
    %
    %
    %
    %
    \hspace{-0.65cm}  
    This work is licensed under a Creative Commons
    Attribution 4.0 International License.
    License details:
    \url{http://creativecommons.org/licenses/by/4.0/}.
}

Natural language understanding (NLU) is impossible without
knowledge about the world.  Large scale knowledge graphs
(KGs) such as Freebase \cite{bollacker2008freebase} are
structures that store world knowledge.  Unfortunately, KGs
suffer from incomplete coverage \cite{min2013distant}; e.g.,
Freebase contains Brandon Lee, but not his
ethnicity.

The knowledge in KGs needs to be expanded to cover more
facts; reasoning is one way to do so.  For example, we
could infer that \emph{(Microsoft, ?, United States)}
instantiates ``CountryOfHQ'' given the  facts
\emph{(Microsoft, IsBasedIn, Seattle)} and \emph{(Seattle,
  LocatedIn, United States)}; or we could infer Brandon Lee's
ethnicity  from his parents' ethnicity,
i.e., answering the query \emph{(Brandon Lee, Ethnicity,
  ?)}\ by  facts \emph{(Brandon Lee, Father, Bruce Lee)}
and \emph{(Bruce Lee, Ethnicity, Chinese)}.  We refer to the
two reasoning examples as ``knowledge base
completion (KBC)'' and ``path query answering (PQA)'',
respectively.

The most successful approach for modeling KGs is the
\emph{embedding
  approach}. It embeds KG elements
(entities and relations) into low-dimensional dense vectors;
controlling the dimensionality of the vector space forces
generalization to new facts
\cite{nickel2011three}.


In this work, we are mainly interested in three issues. (i)
Compared to modeling one-hop KG paths, a bigger challenge is
how to model multi-hop paths, e.g., the path query
\emph{(U.S.A, president$\to$spouse$\to$born\_in, ?)}\ for
the question ``Where was the first lady of the United States
born?'' (ii) How can we address different KG reasoning
problems driven by multi-hop paths in a universal paradigm
rather than via different systems?  (iii) How can we combine
specific multi-hop KG reasoning tasks with generic KG representation
learning, so that KG representation learning can either
stand alone or be incorporated into diverse multi-hop KG-related NLU
problems. We get inspiration from following two types of
work.

First, \emph{prior work in KG reasoning}.
\newcite{guu2015traversing} extend one-hop reasoning regimes
such as TransE \cite{bordes2013translating} to multi-hop
PQA. However, these basic one-hop models do not encode the
relation order when used in compositional training schemes.
For example, path query $q_1$ = (h,
$r_1$$\to$$r_2$$\to$$\cdots$$\to$$r_k$, ?)  will be encoded
into the same embedding as path query $q_2$ = (h,
$r_2$$\to$$r_1$$\to$$\cdots$$\to$$r_k$, ?), resulting in
(often incorrect) prediction of the same tail entity.
Instead, the relation order should influence the prediction.
This limitation in modeling multi-hop relation paths
motivates the \emph{RNN approach}: using recurrent neural
networks (RNN \cite{journalsElman90}) to model relation
paths \cite{neelakantan2015compositional}.
\newcite{das2016chains} further extend this approach by
incorporating entity information and apply it to multi-hop
KBC. Intermediate entities should influence the reasoning
decision. For example, given two paths with the same
relation sequence: (Donald Trump, \emph{child}, Ivanka
Trump, \emph{mother}, Ivana Trump) and (Donald Trump,
\emph{child}, Barron Trump, \emph{mother}, Melania Trump),
even though both paths have the relation sequence [$child$,
  $mother$], the relation between (Donald Trump, Melania
Trump) is ``spouse'' while it does not hold between (Donald
Trump, Ivana Trump) due to the intermediate entities:
``Ivanka Trump'' vs. ``Barron Trump''. Similarly,
paths (JFK, $located\_in$, NYC, $located\_in$, NY) and
(Yankee Stadium, $located\_in$, NYC, $located\_in$, NY)
would predict the same score for target relation
``airport\_serves\_place'' if we do not consider that Yankee
Stadium is not an airport \cite{das2016chains}.

Second, \emph{ sequence labeling tasks} such as
POS tagging, chunking and  NER have been successfully addressed by RNNs
\cite{DBLPcorrHuangXY15,DBLPLampleBSKD16}. These approaches
model the mechanism in a structure of form ``input$_1$,
tag$_1$, input$_2$, tag$_2$, input$_3$, tag$_3$, $\cdots$, input$_t$, tag$_t$'',
which resembles the structure of multi-hop KG paths.

Inspired by this prior work,
we propose \emph{\modelname{}}, Recurrent One-hop
Predictor. Given a head entity,
ROP encodes a multi-hop sequence of relations
and predicts a sequence of entities
using an RNN. More formally,
given relation sequence ``$r_1,
r_2, \cdots, r_t$'' and the head entity $e_h$,
\modelname{}
predicts the
sequence $e_1, e_2, \cdots, e_t$, thus generating a complete
KG path $e_h, r_1, e_1,
r_2, e_2, \cdots, r_t, e_t$.
Intuitively, our model
memorizes history of path context; given a new relation, it
predicts the next entity, then the memory is
updated, and the process -- given new relation, predicting
new entity, updating memory -- keeps going.
Grouping step-wise updates in a chain gives
our model two advantages. (i)  A better vector space
representation of KG entities and relations, with training
signals either from in-path entities or from labels of reasoning tasks or from both. (ii)
A unified approach for two different reasoning
tasks (mh-KBC and mh-PQA) and  state-of-the-art in each.


In summary, our contributions are:
(i) \modelname{}, a novel RNN sequence modeling of multi-hop KG paths that updates entity and relation embeddings
by training signal at each step and leads to better KG embeddings;
(ii)
unified framework for solving different multi-hop reasoning tasks over KGs;
(iii) showing the importance of modeling within-path entities in mh-PQA;
(iv) state-of-the-art results for both mh-KBC and mh-PQA;
(v) releasing an enhanced version of a mh-PQA dataset by adding within-path entities.

\secref{task} introduces the mh-KBC and mh-PQA  tasks. \secref{relatedwork}
discusses related work. \secref{modelname} presents
three \modelname{} architectures and \secref{experiment}
evaluates them. \secref{conclude}
concludes.

\section{Multi-Hop Path Reasoning Tasks}\label{sec:task}
We first give background on the two KG reasoning tasks we address in this work.

\paragraph{Knowledge Base Completion (mh-KBC).}
In mh-KBC, the goal is to  predict new relations between entities
using existing path connections. For example, \emph{(A,
  LivesIn, B)} is implied, with some probability, from
\emph{(A, CEO, X)} and \emph{(X, HQIn, B)}. This kind of
reasoning lets us infer new or missing facts from KGs.

\begin{figure}
\centering
\subfigure[Multi-hop knowledge base completion]{
\label{fig:kbpath}
\includegraphics[height=2in]{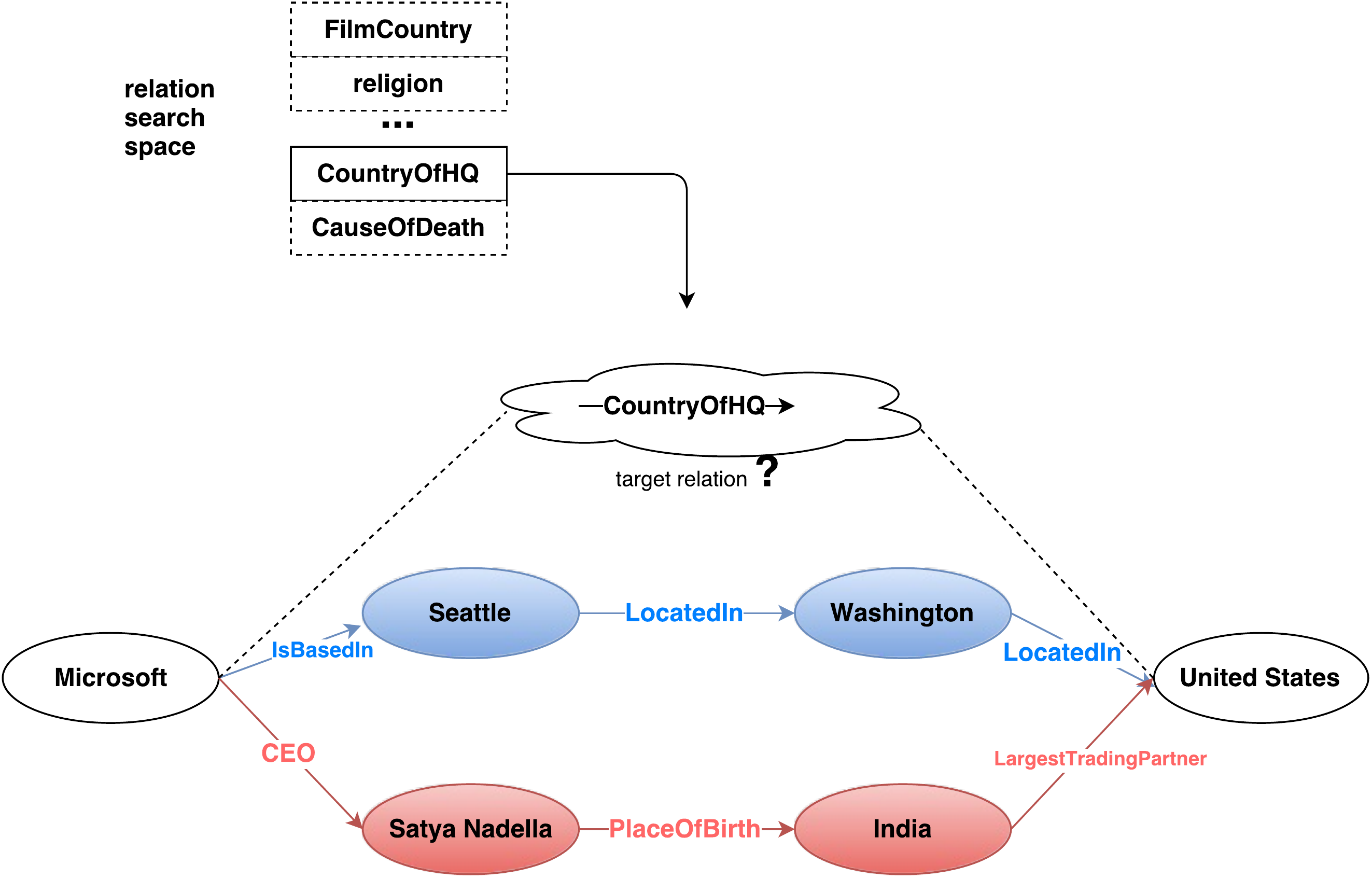}}
\subfigure[Multi-hop path query answering]{
\label{fig:mh-PQA}
\includegraphics[height=1.8in]{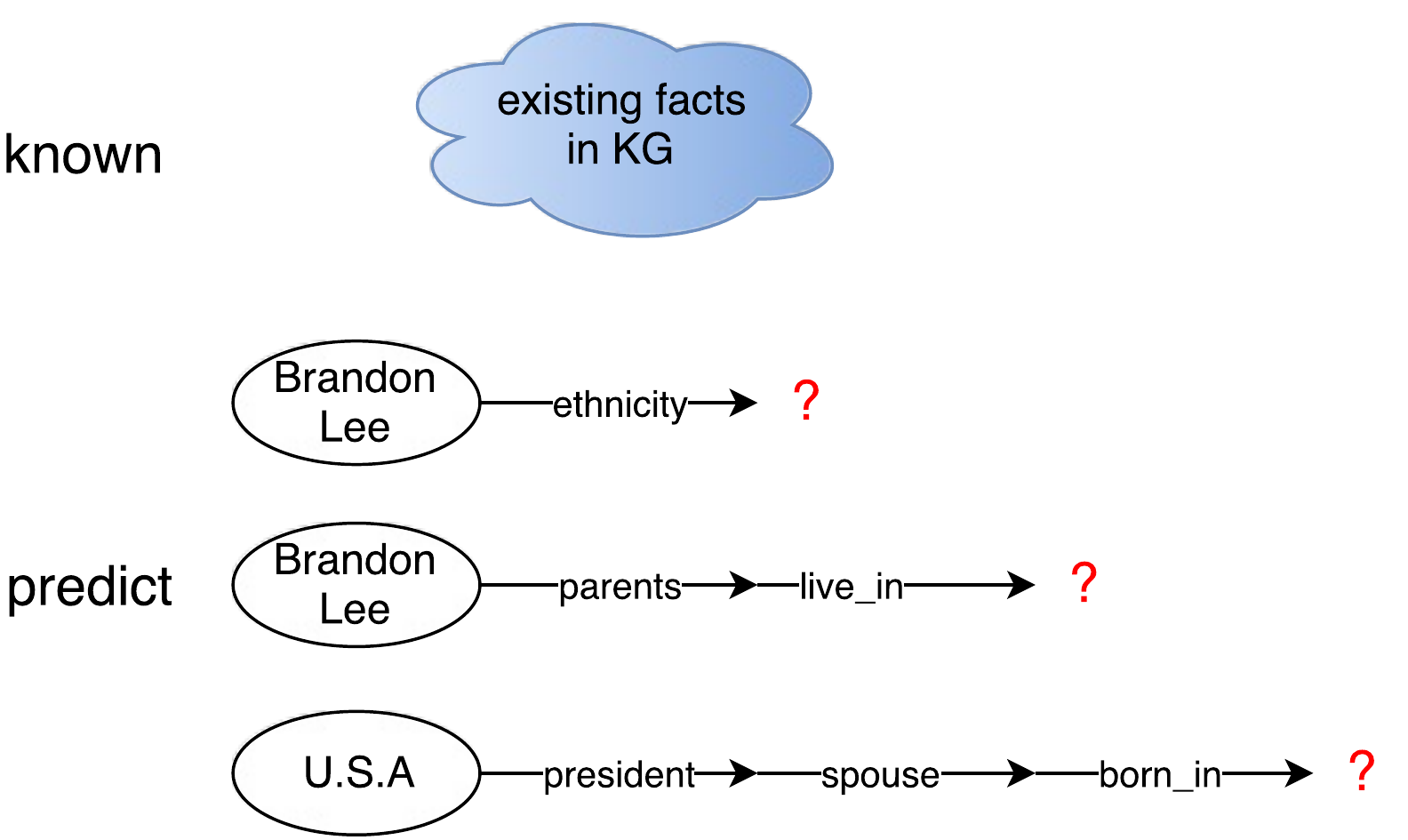}}
\caption{Multi-Hop Path Reasoning Tasks}
\end{figure}



\figref{kbpath} shows two paths between
\emph{Microsoft} and \emph{United States}:
\emph{(Microsoft, IsBasedIn, Seattle, IsLocatedIn,
  Washington, IsLocatedIn, United States)} (blue) and
\emph{(Microsoft, CEO, Satya Nadella, PlaceOfBirth,
  India, LargestTradingPartner, United States)} (red). The task is
then to predict the direct relation that connects
\emph{Microsoft} and \emph{United States}; i.e., \emph{CountryOfHQ} in this case.  There
can exist multiple long paths between two entities; the example shows that the target relation may only be
inferrable from one
path. The difficulty of finding the most informative
path makes this task  challenging.

\paragraph{Path Query Answering (mh-PQA).} In mh-PQA, the goal is to \emph{predict missing properties} of an entity, such as the earlier mentioned  ethnicity of Brandon Lee.
More generally, given an entity and relations of interest, predict what the target entity is, as \figref{mh-PQA} shows. This task corresponds to answering
   compositional natural questions.  For example,
the question
``Where do Brandon Lee's parents live?''
can be formulated by the path query  brandon\_lee/parents/live\_in.
mh-PQA tries to find answers to the path queries and hence compositional questions.
Unfortunately,
KGs often have missing facts (edges), which makes mh-PQA a non-trivial problem.

A path query $q_t$ consists of an initial anchor entity,
$e_h$, followed by a sequence of $t$ relations to be
traversed, $p$ = ($r_1$, $\cdots$, $r_t$). Following \cite{guu2015traversing}, the answer or
denotation of the query  is the set of
all entities that can be reached from $e_h$ by traversing
$p$.

\section{Related Work}\label{sec:relatedwork}

Here we focus on the  multi-hop path reasoning literature. Some work
\cite{neelakantan2015compositional,guu2015traversing,lin2015modeling,Lin0G16,ShenHCG16}
 does some composition over relation paths. Given
relation path $p = (r_1, \cdots, r_t)$, the composition
operation is \emph{add} ($\mathbf{p} =
\mathbf{r}_1+\cdots+\mathbf{r}_t$), \emph{multiplication}
($\mathbf{p} = \mathbf{r}_1\cdots\mathbf{r}_t$) or an
\emph{RNN step}: $\mathbf{p}_i =
\mbox{RNN}(\mathbf{p}_{i-1}, \mathbf{r}_i)$, where $\mathbf{p}_i$
is the accumulated relation information up to step $i$.
Some work explores compositional encoding of long
paths \cite{Lin0G16,ShenHCG16}, but still performs reasoning in one-hop scenario.
\newcite{neelakantan2015compositional} use RNNs to model multi-hop paths.

\newcite{das2016chains} extend the RNN approaches by
leveraging within-path entities into the encoding of inputs
along with relations.  We also include within-path entities,
but we do not give them as inputs; instead, \emph{we force our RNN
to predict them as outputs and do updates at each step
in the path}. This supports representation learning
for KG entities and relations even without task-specific
annotations.  \newcite{ToutanovaLYPQ16} propose a dynamic
programming algorithm to model both relation types and
intermediate entities in the compositional path
representations and test on WordNet and a biomedical KG.
These two works address mh-KBC; for mh-PQA, there is no prior work
on using within-path entities\footnote{\newcite{guu2015traversing}
attempt to measure how severe the cascading errors along the path are by reconstructing the intermediate entities along a path.
Their operation only generates an evaluation score for the
path representation. We instead turn this into a training objective to fine-tune the path representations.}, including   \newcite{das2016chains}, in which  the system uses within-path entities as input, while those entities are not available for testing. So \newcite{das2016chains} use RNN for mh-PQA to encode the relation sequence, but it does not incorporate the intermediate entities involved.


\def\modelnamewidth{5cm}

\section{Recurrent One-Hop Prediction}\label{sec:modelname}

\begin{figure}[h]
\centering
\subfigure[\modelname\_ARC1] { \label{fig:modelname1}
\includegraphics[width=\modelnamewidth]{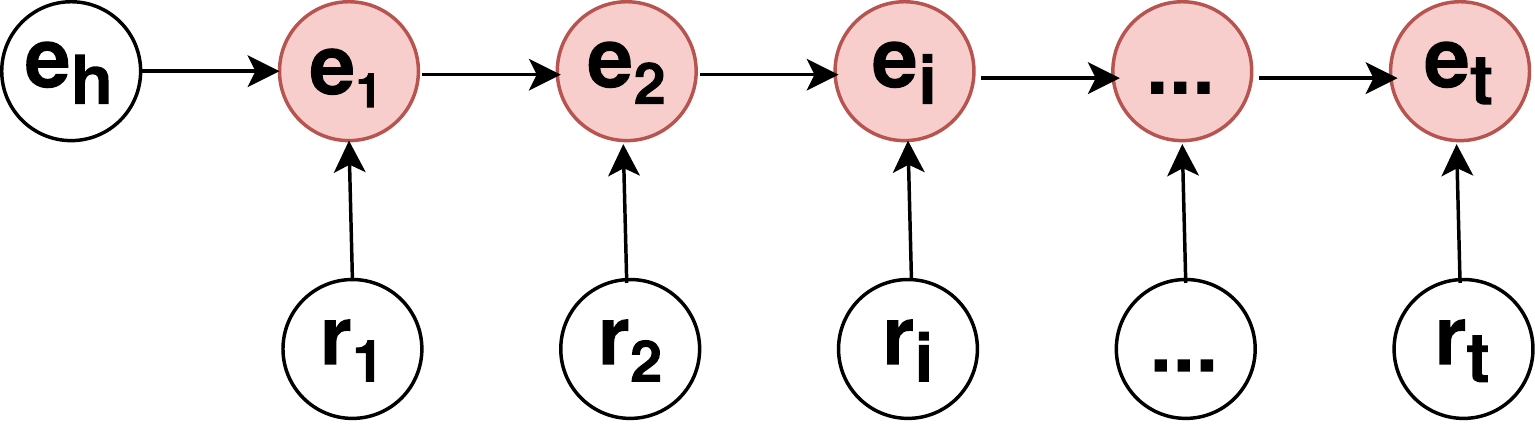}
}
\subfigure[\modelname\_ARC2] { \label{fig:modelname2}
\includegraphics[width=\modelnamewidth]{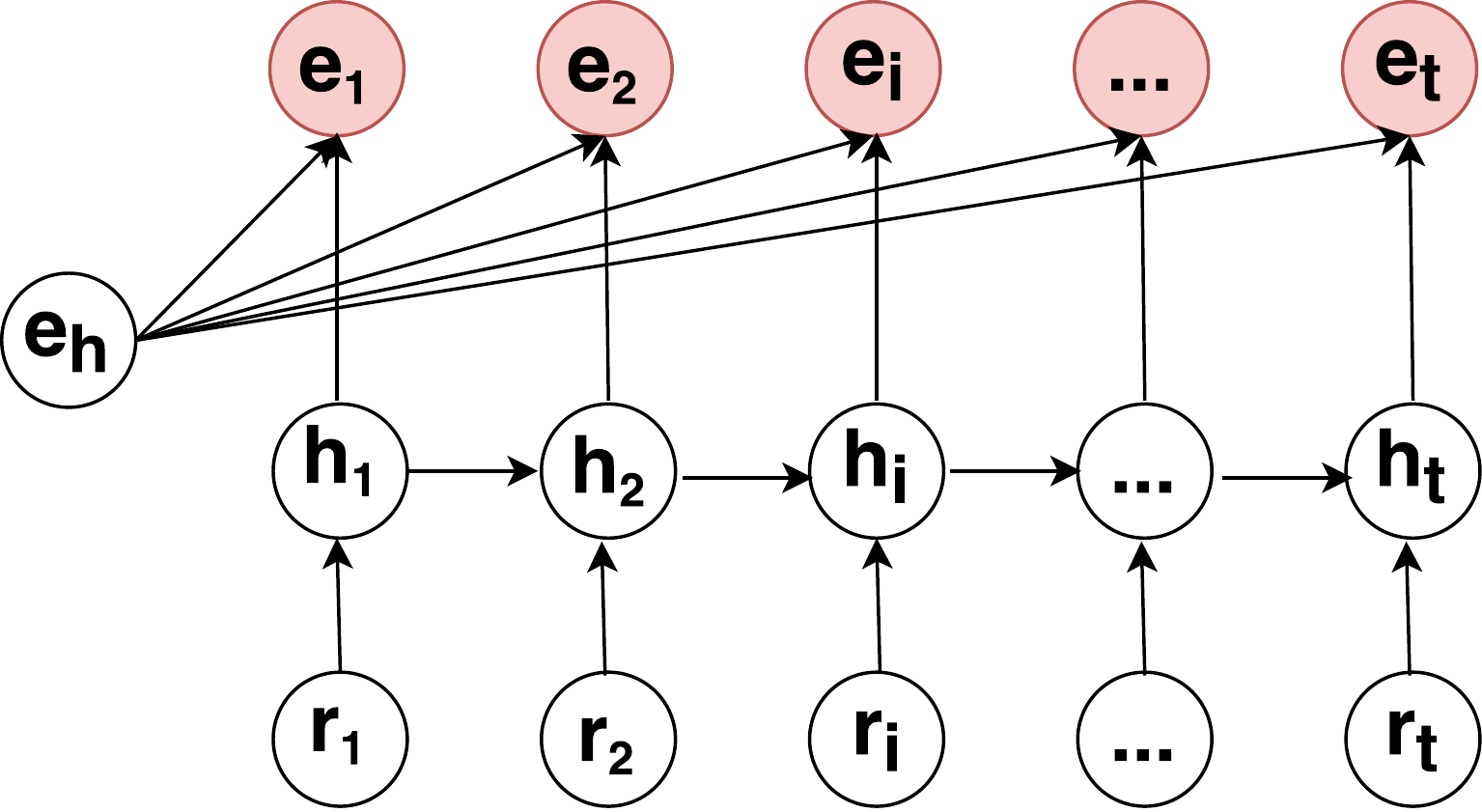}
}
\subfigure[\modelname\_ARC3] { \label{fig:modelname3}
\includegraphics[width=\modelnamewidth]{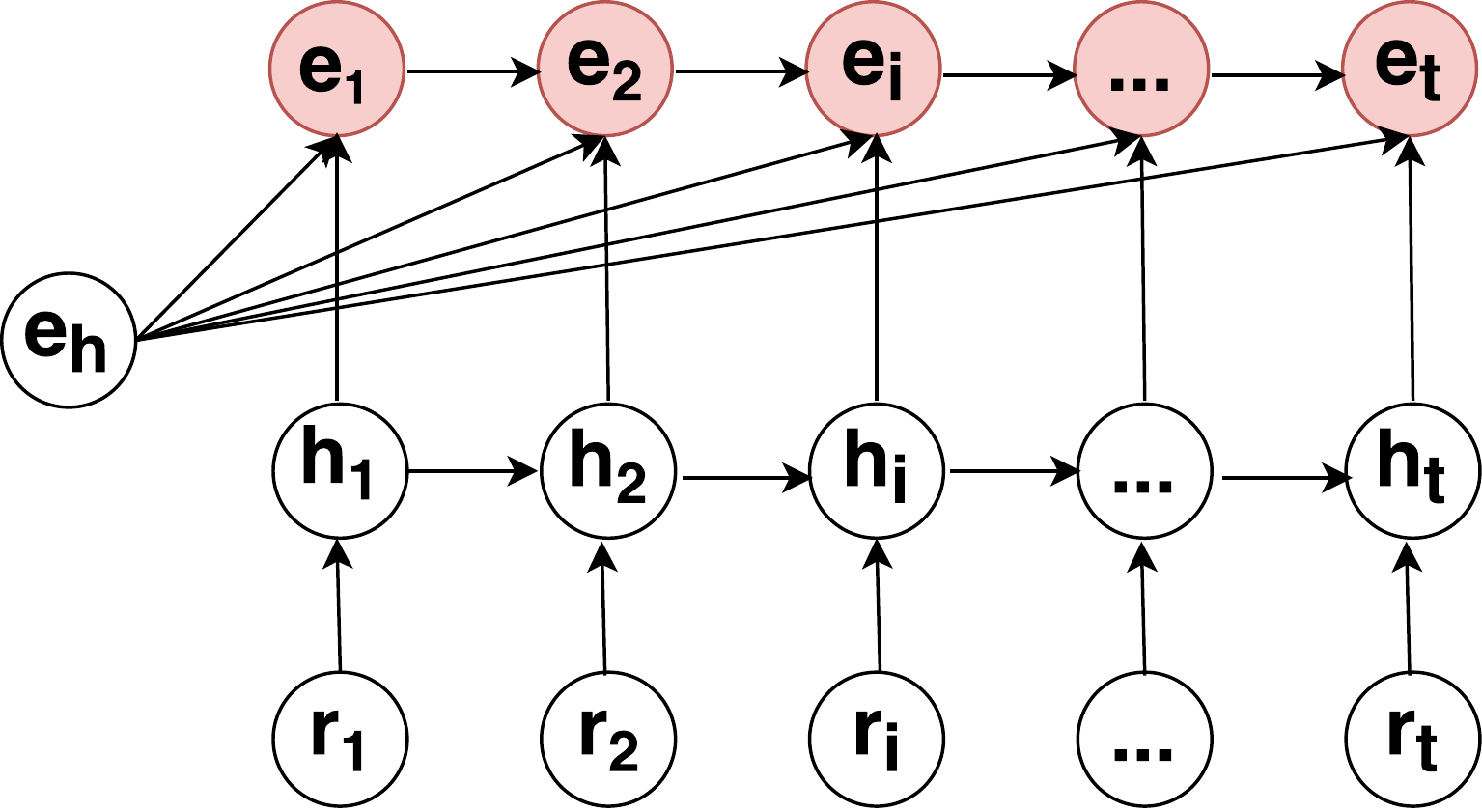}
}
\caption{Three recurrent one-hop predictors}
\label{fig:modelname}
\end{figure}

We propose three ROPs, recurrent one-hop predictors, to model paths such as $p$ = ($e_h$, $r_1$,
$e_1$, $\cdots$, $r_i$, $e_i$, $\cdots$, $r_t$,
$e_t)$.
Entities and relations appear alternately in $p$; $e_h$
and $e_t$ are head  and tail entities;
$t$ steps (hops) connect $e_h$ and $e_t$; each step is a
single fact triple $(e_{i-1}, r_i, e_i)$. We stipulate $\mathbf{e}_i\in\mathbb{R}^{d_e}$ and
$\mathbf{r}_i\in\mathbb{R}^{d_r}$. We allow $d_e \neq d_r$.


Paths are encoded by GRUs \cite{cho2014properties}:
\begin{eqnarray}
\eqlabel{gru1}
\mathbf{g}_z&=&\sigma(\mathbf{x}_i\mathbf{U}^z+\mathbf{h}_{i-1}\mathbf{W}^z)\\
\mathbf{g}_r&=&\sigma(\mathbf{x}_i\mathbf{U}^r+\mathbf{h}_{i-1}\mathbf{W}^r)\\
\hat{\mathbf{h}}_i&=&\mbox{tanh}(\mathbf{x}_i\mathbf{U}^q+(\mathbf{h}_{i-1}\circ \mathbf{g}_r)\mathbf{W}^q)\\
\eqlabel{gru4}
\mathbf{h}_i&=&(1-\mathbf{g}_z)\circ \hat{\mathbf{h}}_i+\mathbf{g}_z\circ \mathbf{h}_{i-1}
\end{eqnarray}
where $x$ is the input sequence with $x_i$ at  position $i$,
$h$ is the output sequence with $h_i$  at position $i$. $g_z$ and $g_r$ are gates. All $\mathbf{U}$s and $\mathbf{W}$s are parameters.
In following, \emph{we define the whole
\eqsref{gru1}{gru4}
as a single GRU step as}:
\begin{eqnarray}\eqlabel{grustep}
h_i = \mbox{GRU}(h_{i-1}, x_i)
\end{eqnarray}
Thus,
we interpret each GRU step as
a composition function of two objects: $h_{i-1}$ and
$x_i$.

We now introduce the three architectures
\modelname\_ARC1, \modelname\_ARC2 and \modelname\_ARC3 that
encode the context ``$e_h$, $r_1$,
$e_1$, $\cdots$, $r_i$'' in different ways to
predict entity $e_i$.

\textbf{\modelname\_ARC1} (\figref{modelname1})
models KG paths as:
\begin{align}
\noindent
\hat{\mathbf{e}}_0= &\mathbf{e}_h\\\noindent
\hat{\mathbf{e}}_i= &\mbox{GRU}(\hat{\mathbf{e}}_{i-1}, \mathbf{r}_i)
\end{align}
$\hat{\mathbf{e}}_0$ is initialized to the embedding of the \emph{true} head entity
$\mathbf{e}_h$. At position $i, i>0$, $\hat{\mathbf{e}}_i$ is the \emph{predicted} entity embedding.

\modelname\_ARC1 is essentially a recurrent process with a pre-set starting state; the key is to use the head entity
embedding $\mathbf{e}_h$ as the initialization of the hidden state of
GRU. We hope this start point guides where the path goes and
what state to reach at each position. As a
result, \emph{relations lie in the input space and entities lie
in the hidden space}. All \emph{predicted} entities
$\hat{\mathbf{e}}_i$ will be compared with the gold intermediate
entities $\mathbf{e}_i$, then the loss (red  in
\figref{modelname}) is used to train the system.

\modelname\_ARC1 only encodes head entity $e_h$ at the
starting hidden state,  possibly far from the tail
entity $e_t$ for long paths. Thus, $e_h$ cannot provide effective
guidance for prediction of $e_t$. This motivates
\modelname\_ARC2, a modification of \modelname\_ARC1.

In \textbf{\modelname\_ARC2}, we want the head entity $e_h$ to participate in the  entity prediction at each step more directly and effectively. To this end,
\modelname\_ARC2 first encodes  the relation sequence as standard GRU:
\begin{equation}
\begin{aligned}\eqlabel{modelname2}
\mathbf{h}_0&=\mathbf{0}\\
\mathbf{h}_i&= \mbox{GRU}(\mathbf{h}_{i-1}, \mathbf{r}_i)
\end{aligned}
\end{equation}
$\mathbf{h}_0$ is initialized to
$\mathbf{0}$.
 $\mathbf{h}_i$,
$i>0$, contains the information of the relation sequence $r_1, r_2, \cdots, r_i$ independent of the head entity $e_h$. To make sure the path reaches the correct state, \modelname\_arc2 then  composes each $h_i$ with $e_h$ to predict $e_i$ as:
\begin{equation}
\eqlabel{comp1}
\hat{\mathbf{e}}_i= \mbox{COMP}_1(\mathbf{e}_h, \mathbf{h}_i)
\end{equation}
As composition function $\mbox{COMP}_1$ we use
ADD (addition, as in TransE) or GRU
(\eqref{grustep}).

In \modelname\_ARC2,  head
entity $e_h$ directly participates in entity prediction at each
step. Unfortunately, there is often another issue
 -- head entity $e_h$ may not match the hidden state
$h_i$ if they are far away from each other -- $h_i$ mainly
encodes some latest relation inputs that are less related to
the head entity. Besides, there is a second information source,
in addition to $e_h$,
that is
clearly relevant for accurate prediction:
the preceding predicted entity $\hat{e}_{i-1}$. But
\modelname\_ARC2 does not make it available.
Our solution is
\modelname\_ARC3. This architecture
predicts the next entity $e_i$ using both
$e_h$ and
 $\hat{e}_{i-1}$, based on the intuition that
$e_i$
is directly related to its predecessor $e_{i-1}$.

\textbf{\modelname\_ARC3} combines the benefits of  \modelname\_ARC1 and \modelname\_ARC2. It encodes the relation sequence
as in \eqref{modelname2} in
\modelname\_ARC2, but composes head entity $e_h$ as well
as the \emph{predicted} entity $\hat{e}_{i-1}$ with $h_i$ to
predict the entity $e_i$ as:
\begin{equation}
\mbox{$\hat{\mathbf{e}}_i= \mbox{COMP}_2(\mathbf{e}_h, \hat{\mathbf{e}}_{i-1}, \mathbf{h}_i)$}\eqlabel{comp2}
\end{equation}
As composition function $\mbox{COMP}_2$, we use
ADD (addition) or an extended GRU
step, defined as:
\begin{eqnarray}
\mathbf{g}_{zh}&=&\sigma(\mathbf{h}_i\mathbf{U}^{zh}+\mathbf{e}_h\mathbf{W}^{zh})
\eqlabel{fancygru1}\\
\mathbf{g}_{rh}&=&\sigma(\mathbf{h}_i\mathbf{U}^{rh}+\mathbf{e}_h\mathbf{W}^{rh})\eqlabel{fancygru2}\\
\mathbf{g}_{zp}&=&\sigma(\mathbf{h}_i\mathbf{U}^{zp}+\hat{\mathbf{e}}_{i-1}\mathbf{W}^{zp})\eqlabel{fancygru3}\\
\mathbf{g}_{rp}&=&\sigma(\mathbf{h}_i\mathbf{U}^{rp}+\hat{\mathbf{e}}_{i-1}\mathbf{W}^{rp})\eqlabel{fancygru4}\\
\hat{\mathbf{h}}_i&=&\mbox{tanh}(\mathbf{h}_i\mathbf{U}^q+(\mathbf{e}_h\circ \mathbf{g}_{rh})\mathbf{W}^{q_h}
+(\hat{\mathbf{e}}_{i-1}\circ \mathbf{g}_{rp})\mathbf{W}^{q_p})\eqlabel{fancygru5}\\
\hat{\mathbf{e}}_i&=&(1-\mathbf{g}_{zh}-\mathbf{g}_{zp})\circ
\hat{\mathbf{h}}_i+\mathbf{g}_{zh}\circ \mathbf{e}_h + \mathbf{g}_{zp}\circ \hat{\mathbf{e}}_{i-1}\eqlabel{fancygru6}
\end{eqnarray}
where super/subscript $h$ refers to \emph{head} entity and $p$ refers to \emph{prior} predicted entity $\hat{e}_{i-1}$.

In following work, we define \eqsref{fancygru1}{fancygru6} as eGRU (extended GRU) step as:
\begin{equation}\eqlabel{h}
\hat{\mathbf{e}}_i = \mbox{eGRU}(\mathbf{e}_h, \hat{\mathbf{e}}_{i-1}, \mathbf{h}_i)
\end{equation}
We extend GRU into eGRU, so that one architecture can compose three objects: a hidden state $h_i$ and two entity states $e_h$ and $\hat{e}_{i-1}$.

\textbf{Training.}
For the gold entity sequence
$e_1, e_2, \cdots, e_t$,
we define  the \textbf{loss function}
as the margin-based ranking criterion used in TransE \cite{bordes2013translating}:
\begin{equation}
\mbox{$l\dnrm{seq} = \sum_i \mbox{max}(0, \alpha\!+\!\mbox{s}(\mathbf{e}_i^-,\hat{\mathbf{e}}_i)\!-\!\mbox{s}(\mathbf{e}_i,\hat{\mathbf{e}}_i))$}\eqlabel{rank}
\end{equation}
where $\alpha$ is the margin,  $e_i^-$  a negative sample
for entity $e_i$ and $\mbox{s}()$ a similarity function.

\textbf{Discussion.}
The three \modelname{} models \emph{differ} in three aspects.

(i)
{\modelname\_ARC1} has only one composition process, i.e., GRU, to encode from head entity $e_h$ to $r_i$;
{\modelname\_ARC2} and {\modelname\_ARC3} each have two composition processes, one composes relation sequence $r_1, \cdots, r_i$ into hidden state $\mathbf{h}_i$, the other composes entities ($e_h$ in {\modelname\_ARC2}, $e_h$ and $\hat{e}_{i-1}$ in {\modelname\_ARC3}) with $\mathbf{h}_i$ to predict $e_i$.

\figref{modelname} shows that {\modelname\_ARC1} uses the GRU hidden states as outputs to compare with gold entities. {\modelname\_ARC2} and {\modelname\_ARC3} instead compose their hidden states with head entity or preceding predicted entity to generate a new output space, then compare with gold entities.

(ii)
{\modelname\_ARC2} only uses the head entity $e_h$ along with the current hidden state $\mathbf{h}_i$ to
predict the next entity $e_i$ whereas
{\modelname\_ARC1} and {\modelname\_ARC3}
in addition use the
{preceding predicted entity} $\hat{e}_{i-1}$. Hence, {\modelname\_ARC3} roughly uses the combined information of {\modelname\_ARC1} and {\modelname\_ARC2} to do the prediction.


(iii)
GRUs like  {\modelname\_ARC2\&ARC3}
often zero-initialize first hidden state $\mathbf{h}_0$.
Setting $\mathbf{h}_0$
to
the head entity
in  {\modelname\_ARC1}
supports prediction of subsequent entities.

The \modelname{} models are \emph{similar} in three aspects.

(i)
They model entities and relations in different spaces: relations  are in the input space, entities are
in hidden or output spaces.
Our motivation is similar to SE,
TransH and TransR (cf.\ \secref{relatedwork}).

(ii)
The gating mechanism  enables flexible
compositions between entities and relations based on path
context. In contrast,
\emph{one-hop KG embedding approaches
model compositions
statically  as shared parameters  and  consider no context.}

(iii) Unlike
\newcite{neelakantan2015compositional},
\newcite{guu2015traversing} and \newcite{lin2015modeling}
(who also do some composition over relation paths),  we
finetune entity/relation
embeddings in each step of the path.
Our intuition is
that many more training signals coming from each step of the sequence enable better learning of
entity/relation embeddings.
\newcite{das2016chains}, as prior work that incorporates entities in the paths,
only update the entity embeddings in the path once, when reaching the end of the path.
We test this effect in our experiments.

\section{Experiments}\label{sec:experiment}
\subsection{Knowledge Base Completion (mh-KBC)}
\seclabel{mh-KBCexp}
\textbf{Dataset.}
We use \newcite{das2016chains}'s dataset, in
which there are 46 query relations, each  has \emph{train, dev and test} files containing
positive and negative entity pairs (head entity, tail entity). Each entity pair is also provided with multiple multi-hop paths connecting them.

This is
a  binary classification task for each query relation: does the query relation hold
between a pair of entities? The classifier
builds a ranked list of entity pairs for corresponding query
relation. Evaluation: mean average precision (MAP) across all 46 relations.


\begin{figure}[t]
\centering
\includegraphics[width=7cm]{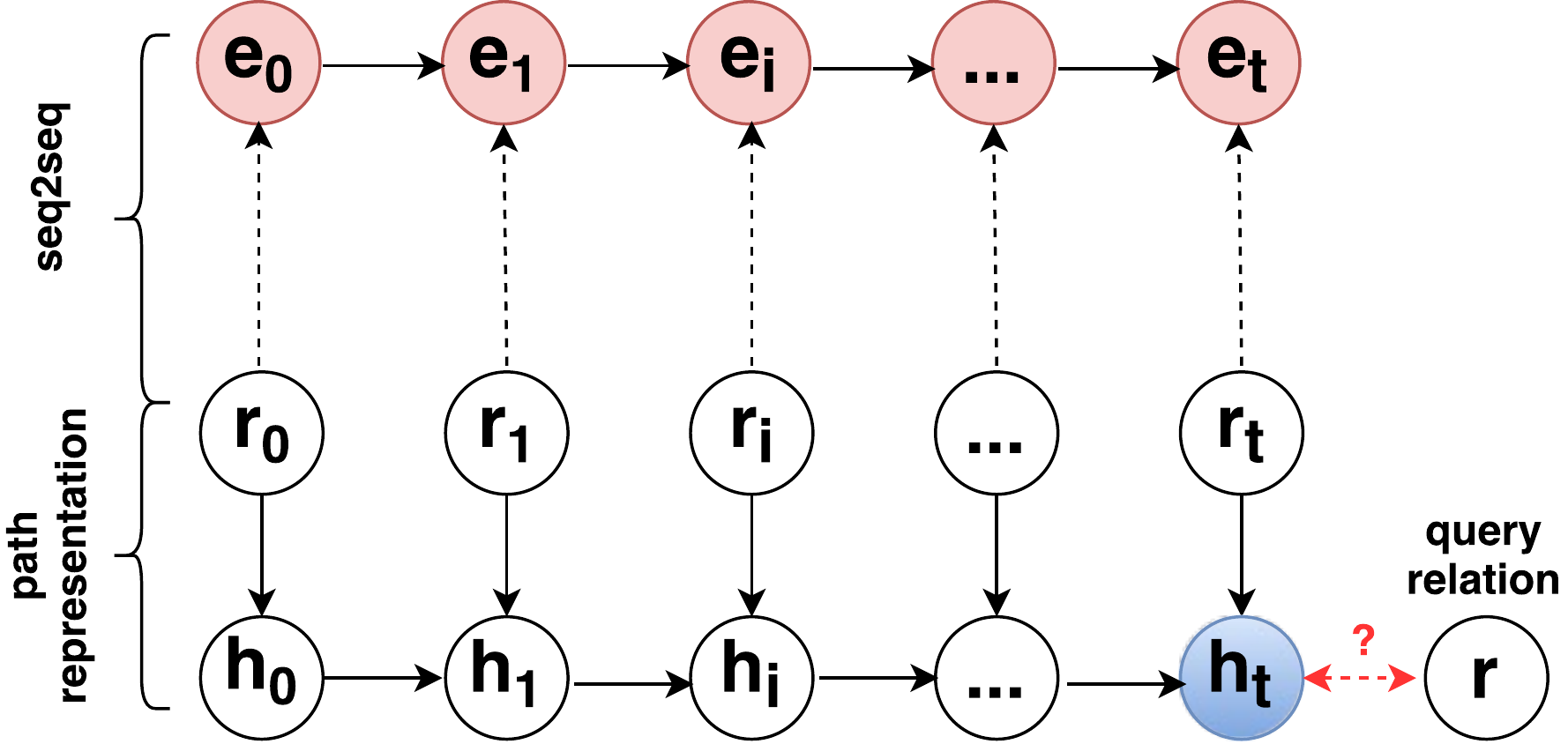}
\caption{Architecture for mh-KBC task} \figlabel{modelnametest}
\end{figure}

\textbf{Task setup.}
We use
\modelname{}  for mh-KBC.
\figref{modelnametest} shows that
we extend
the basic \modelname{} architecture (\secref{modelname})
by a GRU layer
(bottom layer in \figref{modelnametest}), denoted as $\mbox{GRU}_r$, that learns the representation of the relation
sequence. Finally, we use $\mathbf{h}_t$ (the last
hidden state of $\mbox{GRU}_r$, shown in blue
in \figref{modelnametest}) to match the query
relation $\mathbf{r}$.\footnote{Using the last hidden state
  in \modelname{} (which participates in predicting the tail entity) to predict the query relation performed much worse. We suspect the finetuning by tail entity makes it not indicative for query relations.}

Thus, the training loss of this task has two parts:
one comes from the loss of \modelname{} training
($l\dnrm{seq}$, \eqref{rank}); the other is the
prediction loss of query relation, denoted as
$l\dnrm{pred}$.
Our preliminary experiments always showed worse performance of
ADD, so the first loss $l\dnrm{seq}$ here only considers
(e)GRU.
The second loss is as in
\cite{neelakantan2015compositional}; we show that our
\modelname{} part boosts the system as it enables relation paths
to encode entity information with multiple updating -- as
opposed to a single update per path as in \cite{das2016chains}. Similar to $l\dnrm{seq}$, we define:
\begin{equation}
\mbox{$l\dnrm{pred} \!=\! \sum_{j,i} \mbox{max}(0, \beta\!+\!\mbox{s}(\mathbf{h}_{j}^-,\mathbf{r})
\!-\!\mbox{s}(\mathbf{h}_{i},\mathbf{r}))$}\eqlabel{rankmh-KBC}
\end{equation}
where $\mathbf{r}$ is the embedding of a query relation in \figref{modelnametest},
$\mathbf{h}_{i}$ (resp.\ $\mathbf{h}^-_{j}$) is the
representation of positive (resp.\ negative) path
example $i$ (resp.\ $j$),
shown as $\mathbf{h}_t$ in  \figref{modelnametest}.
In testing, all paths  are
ranked in terms of the given query relation.

The target
relation is often not inferrable from all  paths between
head and tail
(see \figref{kbpath}); it can be entailed by a
single path or a subset of paths. Hence, given the
representation of a group of paths, how to match them with
the representation of the target relation is a key
problem. We use max (over all paths)  because we found it works well in
selecting the best path for the target
relation.
See last paragraph of \secref{mh-KBCexp} for discussion.

We use AdaGrad \cite{duchi2011adaptive} with learning rate 0.1. Relation embedding
dimension is 200,  margins $\alpha$ and $\beta$ are 0.5
(\eqsref{rank}{rankmh-KBC}), entity
embedding dimension 200, batch size 20, negative sampling
size 4. Longer paths are
truncated to 8, the first 30 paths
are kept for each entity pair.

\begin{table}[t]
\small
\begin{center}
 \setlength{\tabcolsep}{1mm}
\begin{tabular}{l|l}
Model& MAP \\\hline
PRA \cite{lao2011random} & 64.43\\
PRA+ Bigram \cite{neelakantan2015compositional} & 64.93\\
RNN-path \cite{neelakantan2015compositional} & 68.43\\
RNN-path-entity \cite{das2016chains} & 71.74\\
RNN-path-types \cite{das2016chains} & 73.26\\\hline
\modelname\_ARC1 & 74.23\\
\modelname\_ARC2 & 74.46\\
\modelname\_ARC3 & \textbf{76.16}
\end{tabular}
\end{center}
\caption{Results of mh-KBC task}\label{tab:resulttask2}
\end{table}

\textbf{Results.}
Table \ref{tab:resulttask2} compares our \modelname{} systems
to five baselines.
RNN-path \cite{neelakantan2015compositional} composes
the relations occurring in a path using a vanilla RNN. It
ignores all information about within-path entities and trains separate models per relation.
RNN-path-entity \cite{das2016chains} models the path entities
and improves the results.
RNN-path-type \cite{das2016chains} further improves the result
by representing entities with the sum of their type
embeddings.
Thus, RNN-path-type
uses extra information, the entity types.
Apart from these baselines, it is also feasible to compare with the baselines based on composition of one-hop triple-based embedding models. However, the performance of these baselines is very poor for this task \cite{neelakantan2015compositional} and therefore, we do not include them in our comparison.

\enote{wenpeng}{No, same relation sequences do not necessary means the same target relation. Das' paper gives an example, he argues that without considering the entity types at the two ends, a system could give the same score to a target entity for two paths which have same relation sequence but different entity types.
I guess this is more correct in fine-grained relation classification}

The performance order of our architectures for mh-KBC is {\modelname\_ARC3} $>$ {\modelname\_ARC2} $>$ {\modelname\_ARC1}. All our \modelname{} systems are superior to the
state-of-the-art.
In particular, they are superior to
RNN-path-type \cite{das2016chains}, the state-of-the-art,
even though we do not need and do not use
information about the types of entities.
Comparing \modelname{} performance to RNN-path-entity is more fair,
and this makes it even clearer that our modeling is effective.
\newcite{das2016chains} encode
entities along with relations into the path representation explicitly.  Their
motivation is that entity incorporation prevents prediction of
the same target relation when
relation sequences are the same, but path entities are different.
Our \modelname{} models achieve this implicitly, as the
relation sequence can predict the entity sequence; this
makes sure the representation of the relation sequence is
specific to the entity sequence. As a result, the same goal
can be achieved as \cite{das2016chains}.

In \cite{das2016chains}, an entity and a relation \emph{are
  concatenated as a new unit} in the path; both entity and
relation representations are trained based on the given gold
relation as label.  \modelname{} predicts intermediate
entities and updates the entity/relation embeddings and
other parameters in each step; the training signals can come
from the in-path target entities and from the reasoning
task specific annotations. Thus, \modelname{} makes use
of the in-path structures to learn good quality entity/relation embeddings, which are further  employed and
finetuned to solve the reasoning problem.

This can also explain why max (over all paths) works
well for us while \newcite{das2016chains}
found LogSumExp (over all paths) works better.  As
their system solely relies on  target relations as
training signals, max selection prevents all other
paths from generating gradients to update, so max tends to
select randomly in the initial stage. However, in our
system, the representations of entities and relations get
rich training signals at each path hop, so that the path
representations are more reliable. Hence, max can
select the most informative path and avoid misleading paths
to support the claim of target relation. In a similar
vein, max pooling is widely found more effective
than mean/sum pooling  for classification
as this task mostly relies on the dominant features.

\subsection{Path Query Answering (mh-PQA)}
\textbf{Dataset.} We use the mh-PQA dataset released by
\newcite{guu2015traversing}
and refer to it
as \emph{BaseKGP}\footnote{\newcite{das2016chains} use a dataset based on  the WordNet,
however they conclude that the dataset is not an ideal test
bed for mh-PQA due to some limitations: it is fairly small, with very short paths, few unseen paths during test time, and only one path
between an entity pair. Therefore, we experiment on BaseKGP.}.
BaseKGP contains paths like $e_h,
r_1, r_2, \cdots, r_t, e_t$, where $e_h$ and $e_t$ are
head and tail entities, connected by
relation sequence $r_1, \cdots, r_t$. There are
6,266,058/27,163/109,557 paths in train/dev/test.

BaseKGP is based on
a Freebase subset released by
\newcite{socher2013reasoning}.
The original  subset
contains a collection of Freebase triplets in form of (head,
relation, tail).  \newcite{guu2015traversing}
generated paths by traversing the triplet space.
Paths in BaseKGP of form $e_h$, $r_1$, $r_2$, $\cdots$, $r_t$, $e_t$ do not contain intermediate entities.
We create
\emph{EnhancedKGP} by enhancing each
 BaseKGP path \emph{in train} as follows.
We search entities at each step of a path by
traversing the  subset \cite{socher2013reasoning} until reaching the tail entity $e_t$. When there are
multiple entity choices at a step, we randomly choose
one. EnhancedKGP \emph{train} has the same size as
BaseKGP \emph{train}, except that  paths are filled by intermediate
entities.  Table \ref{tab:enhancedataset} gives statistics. We will release EnhancedKGP, the first dataset for mh-PQA that includes within-path entities.

\begin{table}[t]
\small
\begin{center}
\begin{tabular}{l|rrr}
& \#paths & \#entities & \#relations \\\hline
train & 6,266,058 & 75,043 & 26\\
dev & 27,163 & 41,010 & 26\\
test & 109,557 & 96,858 & 26
\end{tabular}
\quad
\caption{Statistics of EnhancedKGP for mh-PQA}\label{tab:enhancedataset}
\end{center}
\end{table}

\textbf{Task setup.}
We tune parameters on dev. We sample 10 negative
entities for each ground truth entity in the path and use
ranking loss (\eqref{rank}) (with $\alpha$=0.3,
$\mbox{s}()$
= cosine similarity).
For testing, we ignore
intermediate predicted entities and
only output the tail
entity.
We update  parameters
--  relation embeddings, entity embeddings (both
dimension 300) and GRU
parameters --
using AdaGrad with learning rate
0.01 and
mini-batch size 50.

We compare ROP
with the three compositional
training schemes \emph{Bilinear}, \emph{Bilinear-Diag} and
\emph{TransE} in \cite{guu2015traversing}. The compositional
training of TransE, denoted as \emph{Comp-TransE} in this
work, is the state-of-the-art in mh-PQA. For
{\modelname\_ARC2}, we report the performance of using
ADD and GRU for $\mbox{COMP}_1$ in \eqref{comp1}.
Similarly, for
{\modelname\_ARC3}, we report the performance of
using ADD and eGRU
for $\mbox{COMP}_2$ in \eqref{comp2}.

In addition, to better investigate  \modelname{} models, we
run them on both BaseKGP and EnhancedKGP. Note
that since there are no intermediate entities in
{BaseKGP}, {\modelname\_ARC3} is reduced to
{\modelname\_ARC2}. Hence, we report results on
{BaseKGP} only for
{\modelname\_ARC1} and {\modelname\_ARC2}.

Two benchmark metrics are reported for this task \cite{guu2015traversing}: \emph{hits at 10} (H@10),
percentage of ground truth tail entities ranked in the top 10 of all retrieved; and \emph{mean quantile} (MQ), normalized version of mean rank.

\begin{table}[t]
\begin{center}
 \setlength{\tabcolsep}{1mm}
\begin{tabular}{cl|ll}
&methods& MQ & H@10 \\\hline
\multirow{3}{*}{\rotatebox{90}{\scriptsize \begin{tabular}{c}BaseKGP\\ (baselines)\end{tabular}}}
& Comp-Bilinear & 83.5 & 42.1\\
&Comp-Bilinear-Diag & 84.8 & 38.6\\
&Comp-TransE &88.0 & 50.5\\
\hline

\multirow{3}{*}{\rotatebox{90}{\scriptsize \begin{tabular}{c}BaseKGP\\ (our model)\end{tabular}}}
&\modelname\_ARC1 & 88.3 & 52.7\\
&\modelname\_ARC2 (ADD) & 89.3 & 54.3\\
&\modelname\_ARC2 (GRU) & 89.6 & 54.9\\\hline

\multirow{5}{*}{\rotatebox{90}{\scriptsize \begin{tabular}{c}EnhancedKGP\\ (our model)\end{tabular}}}
& \modelname\_ARC1 & 89.4 & 54.2\\\cdashline{2-4}
&\modelname\_ARC2 (ADD) & 90.3 & 55.5\\
&\modelname\_ARC2 (GRU) & 90.5 & 56.3\\\cdashline{2-4}
&\modelname\_ARC3 (ADD)& 90.3 & 55.8\\
&\modelname\_ARC3 (eGRU)& \textbf{90.7} & \textbf{56.7}
\end{tabular}
\end{center}
\caption{Results of mh-PQA}\label{tab:resultsans}
\end{table}

\textbf{Results.}
Table \ref{tab:resultsans} gives results for mh-PQA:
top baselines on BaseKGP (block 1),
\modelname\ results on BaseKGP (block 2) and \modelname{}
results on EnhancedKGP (block 3). Overall, our \modelname{} models all get
new state-of-the-art by large margins (2--4\% H@10 on
BaseKGP, 4--6\% H@10 on EnhancedKGP). The improvements of
the second block over the first are evidence
that recurrent neural networks are an appropriate paradigm
in this task. The third block shows that encoding
intermediate entities in an appropriate way, like
\modelname{} does, can give a further boost.

Comparing {\modelname\_ARC2} and {\modelname\_ARC1}, we
realize that forwarding head entity $e_h$ \emph{directly} to
predict each intermediate entity $e_i$ is better than putting
$e_h$ as the start state of the GRU. We
suspect this is due to the fact that the latest state of
GRU is gradually less influenced by
 $e_h$ when the following relation path is getting longer
and longer. In {\modelname\_ARC2}, no matter how long the
relation sequence $r_1, r_2, \cdots, r_i$ is, we always
compose its whole representation $\mathbf{h}_i$ with the
representation of $e_h$ so that $e_h$ can
participate in the prediction of $e_i$ more directly.

{\modelname\_ARC3}
further improves results by
composing not only head
entity $e_h$, but also the preceding predicted entity
$\hat{e}_{i-1}$ with $h_i$ to predict entity $e_i$. The
result suggests that $\hat{e}_{i-1}$ provides
critical information in this
prediction process. This may be due to the property of RNN
that latest inputs tend to be remembered better than old
inputs, so the latest hidden state $\mathbf{h}_i$ is heavily
influenced by the adjacent relations of position $i$;
employing the preceding predicted entity $\hat{e}_{i-1}$ hence can help the
prediction of $e_i$.

For composition,
GRU/eGRU always surpass ADD -- presumably due to
the higher expressibility of (e)GRU's gating mechanism.

\begin{figure}[t]
\centering
\includegraphics[width=6cm]{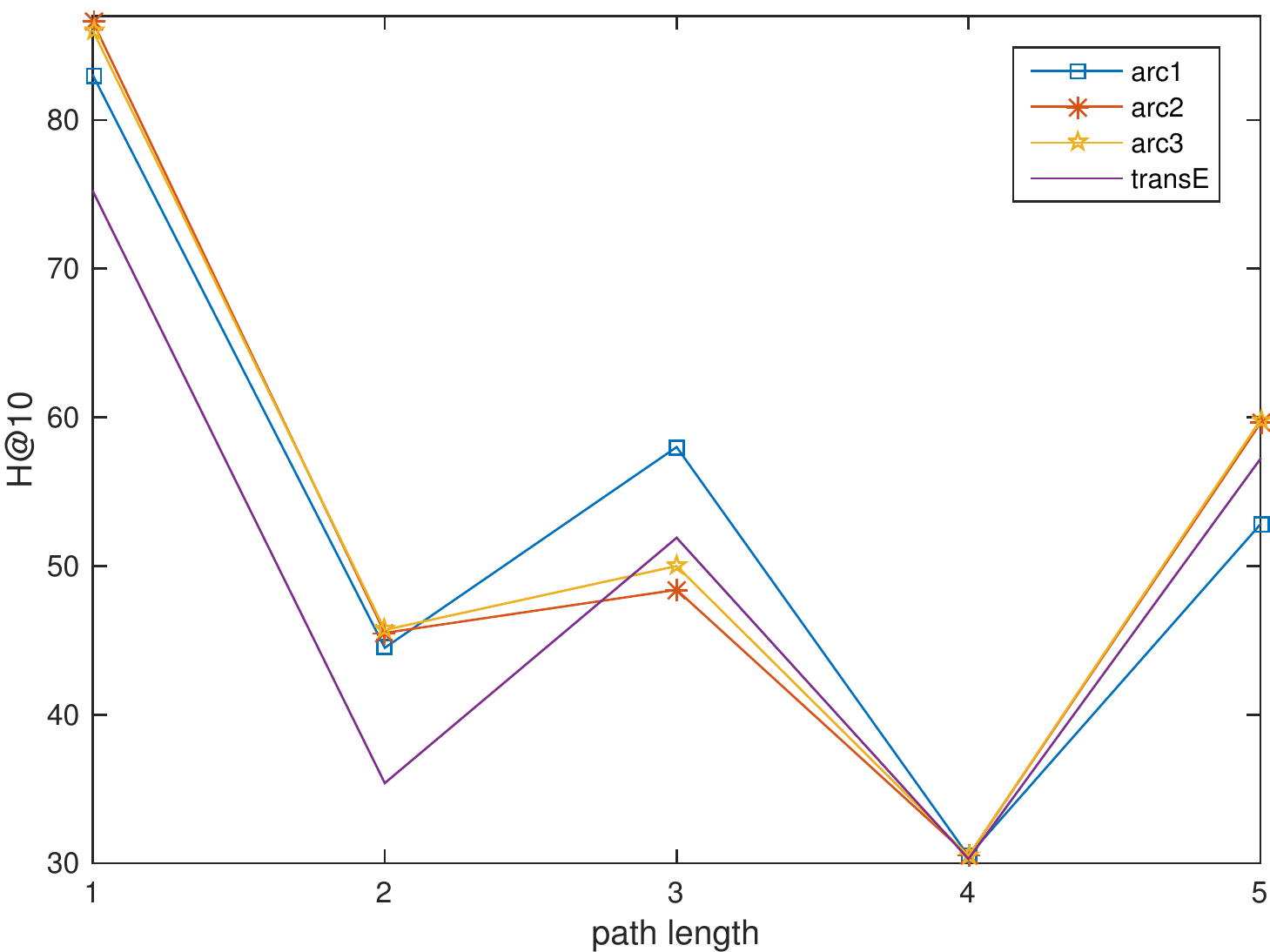}
\caption{H@10 vs. path lengths on test set} \label{fig:length}
\end{figure}

\paragraph{Performance vs. path lengths.} \figref{length} graphs H@10
on different path lengths  for our
three \modelname{} systems and the Comp-TransE baseline.
We expected that
H@10 should decrease gradually. However,
\figref{length} shows that even-length paths are harder than
odd-length ones.
This surprising phenomenon is
\emph{due to a large number of inverse
relations, as most inverse relations  in BaseKGP are 1-to-N connections}.

The percentages of
inverse relations in paths of length 1, 2, 3, 4, 5 are
0.0, 49.7, 38.3, 49.5 and
42.9, respectively. Spearman
correlation coefficients between these percentages and
system performance are always higher than 0.9. Thus, the more
inverse relations, the worse performance.
Unfortunately, to date there is no good way to model pairs of
a relation
 $\mathbf{r}$
and its inverse $*\mathbf{r}$, e.g,
``gender''  and
 ``*gender'', as separate, yet at the same time
as systematically related.
\newcite{guu2015traversing}'s TransE implementation
 enforces $*\mathbf{r}+\mathbf{r}=\mathbf{0}$.
\figref{length} shows this is  not
as effective as expected, perhaps because it fails for
1-to-N projections. \modelname{} treats
inverse relations as independent, but we
did not observe any worse performance. Better modeling
of inverse relations is an interesting challenge for
future work.

\section{Conclusion}\label{sec:conclude}
This work presented three \modelname{} architectures for multi-hop KG
reasoning. \modelname{} models a KG path of
arbitrary length as a pair of a relation sequence and an entity
sequence, using the former to predict the latter by encoding
and decoding step by step.
Our neural sequential modeling of KG paths
enables better learning of entity/relation embeddings
because there are
more training signals at each multi-hop path.
ROPs showed state-of-the-art in two representative
KG reasoning tasks, multi-hop KBC and multi-hop PQA.
Incorporating knowledge from textual sources by
initializing the entity
embeddings with distributional representation of entities \cite{figment15} could improve our results further, which we will explore in the future.

\section*{Acknowledgement}
We gratefully acknowledge the support of
the European Research Council for this research
(ERC Advanced Grant NonSequeToR, \# 740516).

\bibliography{coling2018}
\bibliographystyle{acl}
\end{document}